\documentclass[fleqn,10pt]{wlscirep}

\usepackage[utf8]{inputenc}
\usepackage[T1]{fontenc}
\usepackage{graphicx}
\usepackage{amsmath,amssymb}
\usepackage{booktabs}
\usepackage{url}
\usepackage{xcolor}
\usepackage{tabularx}
\usepackage{longtable}
\usepackage{array}
\usepackage{multirow}
\usepackage{float}
\usepackage{listings}
\usepackage{xcolor}
\usepackage{rotating,booktabs}

\title{
SafeLLM: Extraction as a Hallucination-Resistant Alternative to Rewriting in Safety-Critical Settings
}

\author[1,6,*]{Julia Ive}
\author[1,2,6,$\dagger$]{Felix Jozsa}
\author[1]{Evridiki Georgaki}
\author[3]{Nabeel Sheikh}
\author[3]{Emma Cattell}
\author[4]{Nick Jackson}
\author[1]{Paulina Bondaronek}
\author[2]{Ciaran Scott Hill}
\author[1,5]{Richard Dobson}

\affil[1]{Institute of Health Informatics, University College London, London, UK}
\affil[2]{National Hospital for Neurology and Neurosurgery, Queen Square, London, UK}
\affil[3]{Somerset NHS Foundation Trust, UK}
\affil[4]{King's College Hospital, Denmark Hill, London, UK}
\affil[5]{King's College London, London, UK}
\affil[6]{These authors contributed equally: Julia Ive, Felix Jozsa}

\affil[*]{julia.ive@ucl.ac.uk}
\affil[\dagger]{f.jozsa@ucl.ac.uk}

\begin{abstract}

\textbf{Background:}
Large language models (LLMs) are increasingly used to access organisational documentation, including standard operating procedures (SOPs), HR policies and institutional guidelines. However, retrieval-augmented generation (RAG) approaches that rely on free-form rewriting can introduce hallucinations and unstable trade-offs between completeness and conciseness, particularly in safety- and compliance-critical settings.

\textbf{Objectives:}
To evaluate extraction as a hallucination-resistant alternative to rewriting-based RAG, and to compare strategies that balance precision, recall and safety across document types and model scales.

\textbf{Methods:}
We compare multiple prompting strategies, including line-number-based source selection, extraction of relevant guideline sentences with explicit annotation of safety-related content, and a multi-stage pipeline that refines draft answers using supporting evidence from the source guideline. All approaches outperform direct prompting that relies on the model to copy relevant text from the source document. Experiments are conducted on documents of varying length (local NHS guidelines in acute care and oncology, as well as UK-wide NICE guidelines), and structure using both frontier-scale and smaller locally deployable models. Performance is assessed using automatic metrics and via human expert evaluation of relevance and completeness.

\textbf{Results:}
Line-number selection achieves the strongest results outperforming direct copying and safety-focused strategies across both large and small models, while maintaining high term recall (up to 95\%) and close alignment with source text. Safety-oriented approaches improve precision but introduce systematic omissions, and multi-stage filtering further amplifies this trade-off. Performance varies with document structure: line-based extraction excels in protocol-like content, whereas alternative strategies show advantages in more verbose documents (up to 97\% term recall).

\textbf{Conclusions:}
Line-number selection provides a robust alternative to rewriting-based RAG for accessing organisational documentation. Drafting-based approaches may offer additional benefits in very long-document settings. These findings highlight the importance of adapting retrieval strategies to document structure, model scale and the risks of omission.

\end{abstract}

\begin{document}

\flushbottom
\maketitle
\thispagestyle{empty}

\section*{Introduction}

Administrative and operational documents, including clinical guidelines, standard operating procedures (SOPs), and organisational policies, provide consensus-based instructions for decision-making and task execution \cite{guerra2023clinical}. Across domains, adherence to such documents is a core requirement of good professional practice and is associated with improved safety, compliance, and operational consistency \cite{Scraggs2012,Kresevic2024-se,dean2001decreased,vanwagner2020blood,arnold2009improving,monti2023effect,coba2011resuscitation}.

In practice, high-level guidance is complemented by local or departmental documents that adapt recommendations to specific workflows, constraints, and organisational structures. Users consult these resources to answer focused, task-oriented questions, for example, what actions to take, which approvals are required, or how to escalate a situation. However, their use is often hindered by complexity and time pressure \cite{qumseya2021barriers,tsiga2013influence,freedman2021docs,van2020searching}, challenges that are amplified by increasing workload and growing document volume \cite{torjesen2021doctors,daniels2024perceived,kann2020changes}. In high-stakes or time-critical settings, these limitations can lead to delays, errors, or non-adherence.

Large Language Models (LLMs) offer a promising approach to document-based question answering (Q\&A)~\cite{Singhal2025-ge}, but their deployment is constrained by hallucinations, i.e. generated content not supported by source evidence \cite{Pal2023-ob}. Retrieval-Augmented Generation (RAG) mitigates this by grounding responses in retrieved documents \cite{Lewis2020-qo}, improving factual consistency \cite{Ng2025-be}. However, it does not fully eliminate errors: models may omit critical steps, introduce irrelevant details, or fail to faithfully reproduce source text, posing risks in safety- and compliance-sensitive contexts \cite{Lv2024-qz,Chen2025-xd,Kresevic2024-se,Lewis2025-rm}.

A central challenge in document-based Q\&A is balancing interpretation with faithful reproduction. Some queries require integrating information distributed across multiple sections or adapting conditional instructions to a specific scenario. Others demand precise, verbatim reproduction to preserve meaning, particularly where exact wording carries operational or regulatory significance.

Current approaches focus on improving generation quality, rather than controlling when generation should occur~\cite{Kresevic2024-se,Lewis2025-rm}. Here, we take the opposite approach: we minimise generation. We constrain models to operate directly on source text through extraction, enabling them to identify and present relevant information without rewriting. This reframes the task from answer generation to evidence selection, reducing opportunities for hallucination while preserving fidelity to source material.

Our main contribution is the development and evaluation of an extraction framework for the use case of clinical question answering that minimises hallucinations while balancing the risk of adding information versus omitting clinically critical details. In contrast to rewriting-based approaches, we constrain LLMs to select and copy guideline text, targeting high fidelity to source material, an essential requirement in settings where even small inaccuracies in wording (e.g., drug dosing, contraindications, or escalation criteria) may impact patient safety. We evaluate this approach across multiple guideline use cases that vary in length and structure, including concise acute care protocols from University College London Hospital (UCLH), oncology-focused institutional guidelines from Somerset NHS Trust, and large-scale national recommendations (NICE). These settings capture a spectrum of real-world decision-making scenarios. Evaluation is conducted using human expert-developed question–answer pairs grounded in these documents, reflecting practical clinical information needs under time pressure. We further assess performance across a range of LLM configurations, including both open-weight and commercial models, to characterise robustness across deployment settings. Rather than attempting to make LLMs better generators, we argue that constraining them offers a more reliable path for deployment in safety-critical clinical environments.

\section*{Methods}

\subsection*{Ethical Considerations}

This study was approved by the UCL Review Board as TRE-467283, and approved as a clinical service audit at the National Hospital for Neurology and Neurosurgery, University College London Hospital (UCLH) as 130-202425-SE. The guideline data used in this study contained no patient-level or personally identifiable information. All materials were derived from institutional clinical guidelines intended for professional use and therefore posed no risk to patient privacy.

\subsection*{Data }

We evaluate our methods across three datasets spanning distinct document styles and scales: concise acute care guidelines (UCLH), moderate-length institutional guidelines (Somerset NHS Trust), and large-scale national recommendations (NICE).

The UCLH dataset consists of 96 clinically grounded questions grouped under 24 acute care topics, designed by FJ and NJ to reflect real-world scenarios encountered by frontline clinicians. Topics include, for example, \textit{acute coronary syndrome}, where questions probe escalation decisions (e.g., “If first-line therapy does not relieve symptoms, when is a second dose indicated?”), and \textit{acute kidney injury}, focusing on investigation and management steps (e.g., “Which blood tests are essential on admission?”). Reference answers were constructed via direct selection of relevant guideline lines to ensure grounding in source material.

The Somerset NHS Trust dataset follows a similar question design process but differs in annotation strategy. Here, reference answers were generated exclusively through clinician free writing (NS and FJ). These guidelines are predominantly oncology-oriented and capture acute complications of cancer and its treatment, such as \textit{neutropenic sepsis}, \textit{metabolic disturbances}, and \textit{treatment-related toxicities}. Despite being moderate in length, they are strongly acute in nature, often focusing on time-critical recognition, escalation, and stabilisation decisions in vulnerable patient populations.

In contrast, the NICE dataset was generated synthetically following the protocol of Ding et al.~\cite{Ding2025-ti}, using \texttt{gpt-4o-mini} to produce clinical scenario–question–guideline paragraph triplets from full-length recommendation documents. To reflect realistic model inputs, we merge the scenario description and associated question into a single input prompt. All generated outputs were subsequently manually reviewed by human experts to ensure clinical validity. This dataset captures scenario-related inputs, for example within \textit{chronic kidney disease} or \textit{heart failure} management. A representative example is: 

\textit{“A 68-year-old patient with a history of hypertension and type 2 diabetes presents with progressive fatigue and ankle swelling over several weeks. Blood tests show reduced eGFR and elevated creatinine. According to current guidelines, what investigations and initial management steps should be undertaken, and when should specialist referral be considered?”}

This contrasts with the shorter, focused queries in UCLH and Somerset, such as: \textit{“Which blood tests are essential in acute kidney injury?”} (see Tables~\ref{tab:task_questions_long} and \ref{tab:oncology_questions}).

All underlying guideline documents were extracted from PDF files using the \texttt{pdftotext} utility to preserve original wording, followed by manual cleaning to remove tables of content, correct formatting artefacts and ensure fidelity to source content.

Across datasets, question formulation varies substantially, from concise queries in UCLH (mean 9.6 words) to longer, scenario-based inputs in NICE (mean 45.9 words), with Somerset NHS Trust occupying an intermediate position. Importantly, the guideline corpora differ by an order of magnitude in length: short UCLH documents (mean 746 words), moderate-length Somerset NHS Trust guidelines (mean 3,153 words), and long-form NICE recommendations (mean 5,224 words; maximum 8,007 words) (Table~\ref{tab:data_stats}). 

\begin{table}[t]
\centering
\footnotesize
\setlength{\tabcolsep}{4pt}
\renewcommand{\arraystretch}{1.1}
\caption{
Summary statistics of the evaluation datasets spanning three document-length regimes: concise acute care guidelines (UCLH), moderate-length institutional guidelines (Somerset NHS Trust), and large-scale national recommendations (NICE). Values are reported as mean (min--max).
}
\label{tab:data_stats}
\begin{tabular}{lccc}
\toprule
 & \textbf{UCLH} & \textbf{Somerset NHS Trust} & \textbf{NICE} \\
\midrule
\textbf{Questions (n)} & 96 & 50 & 500 \\
\textbf{Guidelines (n)} & 25 & 6 & 10 \\
\midrule
\textbf{Question length (words)} 
& 9.6 (3--17) 
& 11.2 (2--21) 
& 45.9 (28--78) \\
\textbf{Answer length (words)} 
& 39.4 (1--139) 
& 36.5 (5--145) 
& 28.0 (9--87) \\
\midrule
\textbf{Guideline length (lines)} 
& 73 (34--152) 
& 438 (317--636) 
& 881 (258--1466) \\
\textbf{Guideline length (words)} 
& 746 (375--1360) 
& 3153 (2372--4510) 
& 5224 (1437--8007) \\
\midrule
\textbf{Total corpus (words)} & 18{,}578 & 21{,}025 & 52{,}236 \\
\textbf{Total corpus (lines)} & 1{,}824 & 2{,}825 & 8{,}809 \\
\bottomrule
\end{tabular}
\end{table}

\subsection*{Prompting Strategies for Guideline-Based Clinical Question Answering}

\begin{figure}[t]
\centering
\includegraphics[width=0.7\linewidth]{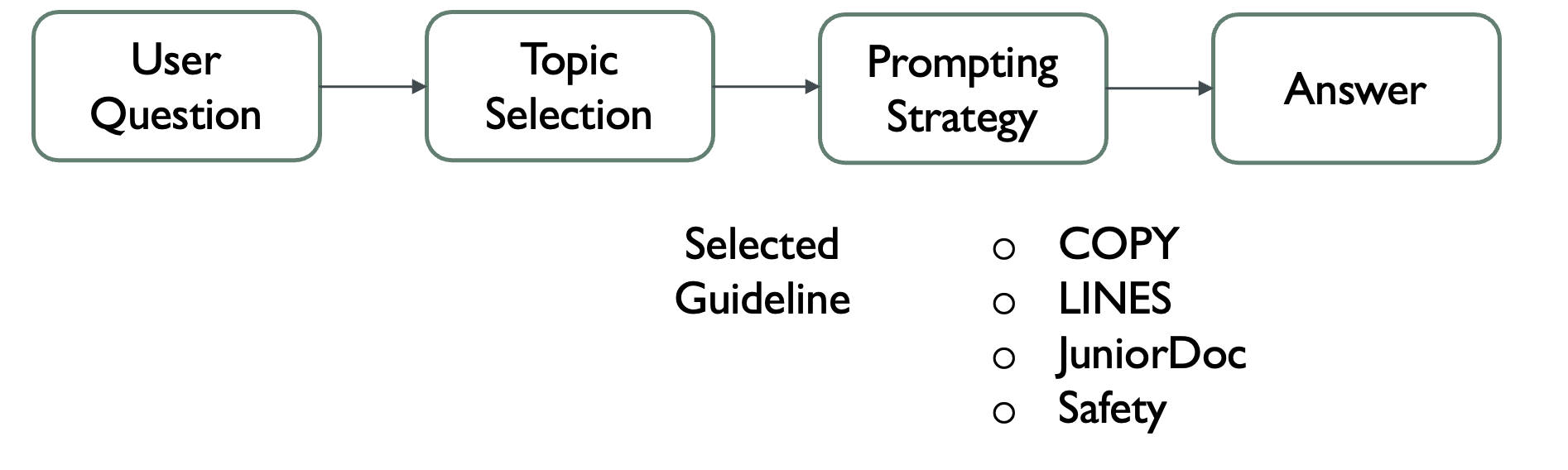}
\caption{
Overview of the proposed guideline question-answering pipeline.
Given a user question, the system first performs \textbf{Topic Selection}
to identify the most relevant clinical guideline from the guideline index.
Once the relevant guideline is selected, the system applies one of several
\textbf{prompting strategies} to extract the evidence required to answer the
question. The evaluated strategies include:
(i) \textbf{COPY}, where the model extracts verbatim sentences;
(ii) \textbf{LINES}, where the model selects relevant guideline line
numbers;
(iii) \textbf{JuniorDoc}, a multi-agent workflow where a junior-doctor-style
draft answer is generated, verified against the guideline, and filtered by a
safety editor; and
(iv) \textbf{Safety}, which extracts the guideline sentences relevant to the question and annotates safety-critical content to support safe evidence-based responses.}
\label{fig:pipeline_prompting}
\end{figure}

This section outlines the methods evaluated in this study, starting with a baseline retrieval approach based on semantic similarity between the user question and guideline text. We then present a set of prompting pipeline methods that combine an initial topic identification step with different prompting regimes for extracting relevant evidence from clinical guidelines (see Fig.~\ref{fig:pipeline_prompting}, see summary of the approaches in Table~\ref{tab:prompting_strategies}).

\paragraph{Baseline Retrieval (BASE)}

As a baseline retrieval strategy, we construct a repository of information chunks derived from guideline text, each representing a self-contained unit of clinically relevant content that may be used to answer a question. From each guideline, we extract guideline sections and, for each section, the section title, section text, and a representative actionable question.

These sections are extracted automatically from guideline paragraphs using an LLM prompt. This section extraction prompt (Listing~\ref{lst:baseline-subparts}) decomposes each paragraph into smaller, clinically meaningful overlapping subparts. For each section, it then generates a short descriptive title and an actionable clinical question. By associating each text segment with a representative question, we aim to reduce sensitivity to phrasing variation between guideline text and associated queries.

\begin{figure}[t]
\centering
\includegraphics[width=0.8\linewidth]{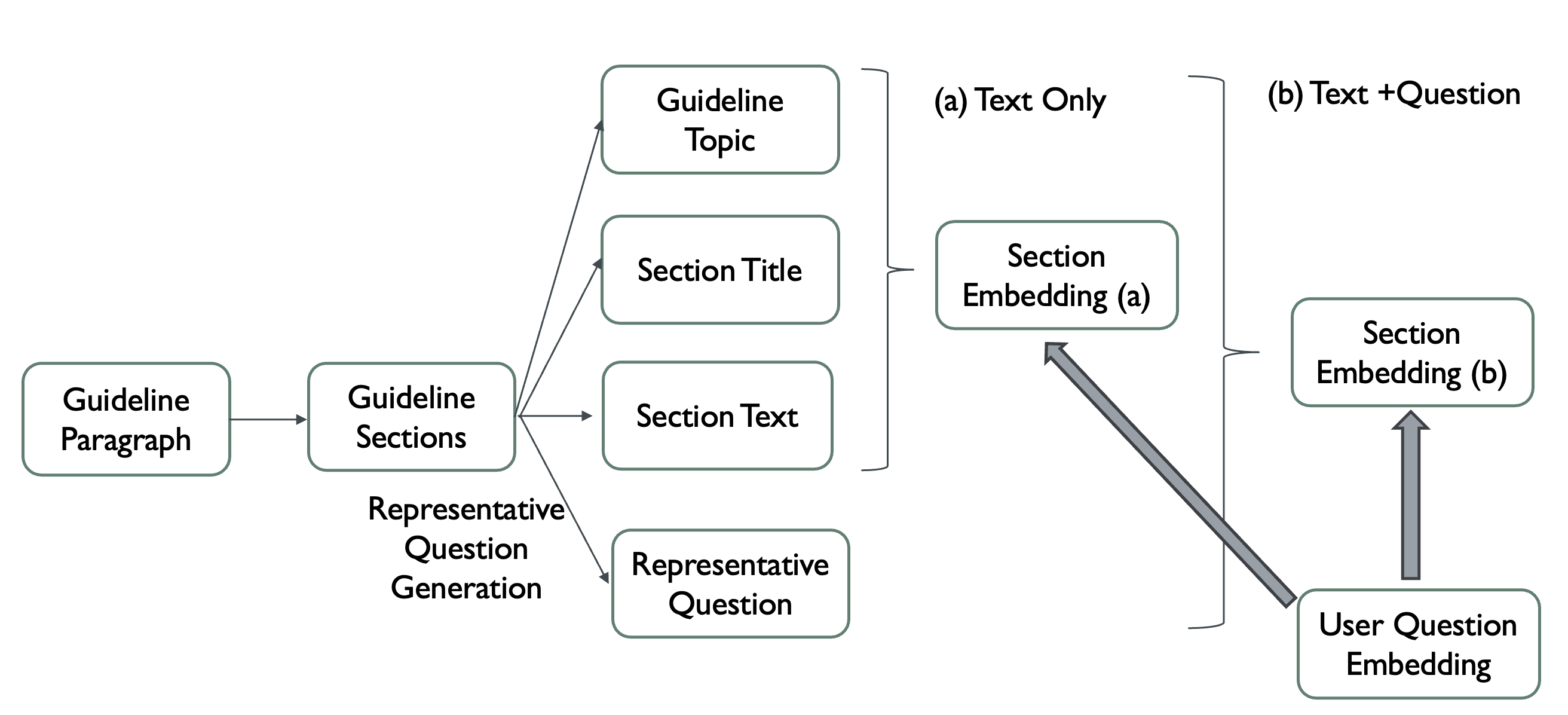}
\caption{
Construction of the representative question–section index used by the baseline
retrieval strategies (BASE). Each guideline document is first segmented into
paragraphs, from which clinically coherent overlapping sections are extracted.
For every section, a representative question and section title are generated using an LLM. Each subpart is represented by four elements: the guideline topic, section title, representative question, and verbatim section text. Two indexing variants are considered:
(a) \textit{text}, where embeddings are computed from guideline topic, section title and text concatenated together into a single string; and (b) \textit{text + question}, where embeddings are computed as in (a) plus the representative question. Representative questions act as a semantic interface between guideline text and user queries. At query time, the user question is embedded using the same embedding model and
compared against section embeddings via cosine similarity.}
\label{fig:baseline_index_pipeline}
\end{figure}

At indexing time, each subpart is embedded using one of two representations (Figure~\ref{fig:baseline_index_pipeline}): (a) \textit{text}, based on guideline topic, section title, and section text concatenated into single string; or (b) \textit{text + question}, which additionally incorporates the representative question (guideline topic, section title, section text and representative question concatenated into single string). At query time, the user question is embedded using the same model and compared to all embeddings (variant (a) or (b)) via cosine similarity. Retrieval is then performed by selecting the subparts whose representations are most semantically aligned with the query.

\begin{table}[H]
\centering
\footnotesize
\caption{Summary of prompting strategies. All methods begin with Topic Selection, which identifies the most relevant guideline document before extraction.}
\label{tab:prompting_strategies}
\begin{tabular}{p{2.5cm}p{3cm}p{6cm}p{3cm}}
\hline
\textbf{Abbrev} & \textbf{Strategy} & \textbf{Key Idea} & \textbf{Hallucination Control} \\
\hline

BASE & Baseline Retrieval & Retrieve relevant guideline sections using embedding similarity between user queries and those sections & Retrieval-only baseline \\

Topic & Topic Selection & Select the single most relevant guideline document before answering the question. & Limits search space to one guideline \\

COPY & Verbatim Sentence Extraction & Model copies only exact sentences from the guideline required to answer the question & Enforces copying \\

LINES & Line Selection & Model outputs line numbers corresponding to relevant guideline text. The final answer is reconstructed from those lines & Guarantees verbatim text. Eliminates hallucination \\

Junior & Junior Doctor Workflow & Multi-agent reasoning: junior doctor draft answers are rewritten using supporting evidence from the source guideline & Grounding \\


Safety & Annotated Safety Extraction & Model selects the relevant set of guideline sentences that answer the question and annotated each with a role label (e.g., answer, detail, safety) & Copying with explicit role labelling \\

\hline
\end{tabular}
\end{table}

\paragraph{Step 1: Topic Selection (Topic)}

Topic selection is the first step across all strategies. Given a user query, the model selects exactly one guideline topic from a predefined index of topics. This design follows the workflow observed in clinical practice. Questions typically relate to a single diagnosis, meaning clinicians first identify the relevant guideline before consulting detailed sections (see Listing~\ref{lst:topic-selection}).

Since our study focuses on acute medical problems encountered on an undifferentiated hospital ward, clinical questions can generally be mapped to a single primary issue, even for patients with substantial multimorbidity. Furthermore, many acute conditions, such as hyperkalaemia or hypoglycaemia, are managed according to standard protocols that are largely independent of a patient's underlying comorbidities. Where management does require adaptation for specific patient groups, such as those with chronic kidney disease, these considerations are typically captured within dedicated sections of the same guideline. This structure supports a topic-first retrieval strategy, in which a single guideline is identified before relevant subsections are selected.

\paragraph{Step 2: Evidence Retrieval Strategies}

\paragraph{Verbatim Sentence Strategy (COPY)}

We first introduce \textsc{COPY} as a prompting baseline that probes the model’s ability to directly copy relevant information from the guideline. The model is instructed to return only verbatim sentences that answer the user question, with explicit constraints forbidding paraphrasing and requiring exact reproduction of the source text. 

This setup probes models' capacity for faithful extraction, providing a reference point for more constrained strategies (see Listing~\ref{lst:copy}).

\paragraph{Line Selection Strategy (LINES)}

Building on this baseline, the \textsc{LINES} strategy introduces a more sophisticated mechanism. The guideline text is augmented with line numbers and provided to the model together with the user question. The model is then tasked with identifying the minimal set of line numbers required to answer the question.

The final answer is reconstructed deterministically by retrieving the corresponding guideline lines in post-processing. This approach enforces tighter control over the extracted content and guarantees the absence of hallucinated text, as the output consists exclusively of selected guideline line numbers (see Listing~\ref{lst:line-selection}).

\paragraph{Junior Doctor Workflow (Junior)}

We also design a multi-agent prompting strategy that mimics a realistic clinical reasoning workflow.

\begin{figure}[H]
\centering
\includegraphics[width=0.7\linewidth]{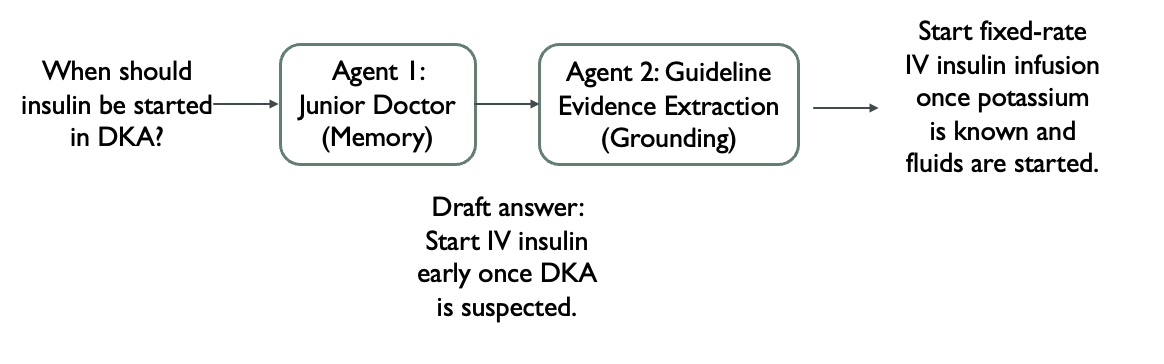}
\caption{
Illustration of the \textbf{JuniorDoc} prompting workflow. The user question is first processed by \textbf{Agent~1}, which acts as a junior clinician reasoning from memory and produces an initial draft answer. This draft may be incomplete or imprecise because it did not yet consult the guideline text. \textbf{Agent~2}
then grounds the answer by extracting verbatim sentences from the relevant
clinical guideline section that support, refine, or correct the initial draft.
This multi-stage design combines
clinical reasoning and evidence grounding.
}
\label{fig:juniordoc_workflow}
\end{figure}

Agent~1 acts as a junior doctor and generates an initial answer based only on training memory, without consulting external documents. This simulates the real-world setting in which clinicians often rely on prior knowledge under time pressure. The agent is instructed to provide concise answers containing actionable details such as thresholds, medication doses, or monitoring steps (see Listing~\ref{lst:juniordoc-agent1}). Agent~2 then compares the draft answer against the official guideline text and selects the guideline sentences that best support, clarify, or correct it, preserving their original order (see Listing~\ref{lst:juniordoc-agent2}, see Figure~\ref{fig:juniordoc_workflow}).


\paragraph{Minimal Safety Extraction (Safety)}

We also evaluate a prompting strategy that instructs the model to extract the minimal safe set of sentences required to answer the user’s question. The model is prompted to select the smallest subset of full sentences that (i) directly answer the question and (ii) include any safety-critical information. Each selected sentence is annotated with a single reason label indicating its role in the response: \texttt{answers\_question}, \texttt{adds\_required\_detail}, or \texttt{safety\_critical}. This prompt also enforces strict constraints: sentences must be copied verbatim from the guideline, paraphrasing is not allowed, and the original order must be preserved.  

The Safety strategy is motivated by the observation that guideline answers often contain both essential recommendations and surrounding explanatory text. By explicitly defining roles, the model is guided to retain only the most relevant content.

\subsection*{LLM Types}

To evaluate the robustness of our methods across different model architectures and deployment settings, we conducted experiments using several open-weight LLMs available through the Ollama framework\footnote{\url{https://ollama.com}}. Specifically, we evaluated \texttt{gpt-oss-120b}~\cite{OpenAI2025-bh} and \texttt{qwen3:32b}~\cite{Yang2025-ii}. These models provide openly available weights, enabling reproducibility and supporting deployment in secure environments where external API access may be restricted (e.g., clinical or trusted research environments). Using open-weight models also allows evaluation under controlled and identical inference conditions. All models were run with default hyperparameters except for the temperature parameter, which was fixed at 0 to minimise non-deterministic variation.

For the baseline retrieval strategy (BASE), we additionally experimented with three embedding models available through Ollama: \texttt{nomic-embed-text}, \texttt{embeddinggemma}, and \texttt{qwen3-embedding}. These models differ primarily in their training objectives and underlying model families. \texttt{nomic-embed-text} is designed specifically for semantic search and retrieval tasks, \texttt{embeddinggemma} provides lightweight embeddings derived from the Gemma model family \cite{Gemma-Team2024-mq}, and \texttt{qwen3-embedding} is based on the Qwen architecture and trained to produce dense representations aligned with instruction-tuned models.

For the open-access NICE guideline data, we additionally evaluate large-scale API-based models to assess how our extraction strategies generalise to frontier models beyond locally deployable setups. Specifically, via the Replicate platform, we include models spanning a wide scale range: \texttt{deepseek-ai/deepseek-v3} (very large, ~671B parameters), \texttt{qwen/qwen3-235b-a22b-instruct-2507} (medium, 235B parameters), and \texttt{gpt-5o-mini} (compact, parameter count undisclosed, typically <30B). We deliberately exclude \texttt{gpt-4o-mini} from this evaluation, as it was involved in the creation of the synthetic NICE dataset. 

\subsection*{Evaluation}

We evaluate the proposed methods using both automatic metrics and human assessment.

\paragraph{Automatic Evaluation}

We first report average answer length (AnsLen) to understand model verbosity, as longer outputs tend to increase recall-oriented metrics while potentially reducing precision.

We then evaluate overlap with reference answers using ROUGE-L~\cite{lin2004rouge}, which can be interpreted simply as a form of \textit{sentence-level recall}: it measures how much of the reference content is recovered \textit{in the correct order}, rewarding correct reconstruction of long chunks of text rather than fragmented matches.

To capture more fragmented lexical overlap, we report word-level precision, recall, and F1 (P, R, F1). Both reference and generated answers are normalised by removing non-alphanumeric characters, lowercasing, and splitting into tokens on space. Precision is defined as the proportion of generated tokens supported by the reference, recall as the proportion of reference tokens recovered, and F1 as their harmonic mean. This formulation is robust to word order and minor formatting differences.

Finally, we assess entity-level fidelity using medical named entity recognition (NER, medical term) precision, recall, and F1. 

We use MedCAT \cite{Kraljevic2021-ln} for biomedical named entity recognition and concept linking. Specifically, we employ the latest available MedCAT model packs based on SNOMED CT and UMLS, accessed through UMLS-authenticated releases.\footnote{~\url{https://medcat.readthedocs.io/en/latest/main.html}} MedCAT identifies clinical concepts in guideline texts and maps them to standard biomedical terminologies. Evaluation is performed over sets of unique extracted term spans, with precision and recall defined as the overlap between predicted and reference entity sets, and F1 as their harmonic mean. This provides a targeted measure of whether clinically salient concepts (e.g., drugs, conditions, symptoms) are correctly preserved, independent of surrounding text.

Together, these metrics provide complementary views of extraction quality: AnsLen captures verbosity, ROUGE-L reflects sentence-level overlap, word-level scores capture fragmented lexical overlap, and medical term metrics focus specifically on clinically relevant entities.

\begin{table}[t]
\centering
\footnotesize
\caption{Automatic evaluation metrics highlighting complementary aspects of verbosity, coverage and clinical fidelity.}
\label{tab:metrics}
\begin{tabular}{p{2.8cm}p{3.2cm}p{6.2cm}}
\hline
\textbf{Metric} & \textbf{Type} & \textbf{Purpose in this study} \\
\hline

AnsLen & Output length & Measures average answer length; captures verbosity \\

ROUGE-L & Sequence-level overlap & Measures ordered overlap with the reference (longest common subsequence); captures sentence-level recall and rewards coherent reconstruction \\

Word P/R/F1 & Word-level precision, recall and F1 & Word overlap robust to formatting differences. Precision reflects unsupported additions, recall reflects coverage, and F1 balances both \\

MedicalTerm P/R/F1 & Medical term overlap & Evaluates preservation of clinically salient concepts \\

\hline
\end{tabular}
\end{table}

\paragraph{Human Evaluation}

In addition to automatic metrics, we conducted a human evaluation to assess the clinical quality of the generated answers. While automatic metrics quantify textual overlap, they are limited in their ability to capture clinical correctness, contextual appropriateness, and the safety implications of missing or added information. In guideline-based question answering, small deviations in wording, omission of key recommendations, or inclusion of irrelevant details may not substantially affect metric scores but can have significant clinical consequences.

Three authors (FJ, EG, and NS; only one of those authors participated in the UCLH guideline data creation) independently reviewed the outputs while having access to the original guideline PDF documents used as source material. Importantly, the evaluators did not have access to the reference answers used in the automatic evaluation. They assessed each system output directly against the guideline documents.

The evaluators scored answers along two dimensions. The first criterion was \textit{correctness}, indicating whether the information provided was relevant to the clinical question (categories: \textit{very relevant}, \textit{relevant}, \textit{irrelevant}). The second criterion was \textit{completeness}, indicating whether the answer adequately covered the relevant guideline content (categories: \textit{minor omission}, \textit{major omission}, \textit{just right}, \textit{minor addition}, \textit{major addition}). Major omissions are particularly important in clinical contexts because missing key recommendations may lead to incomplete or unsafe guidance.

\section*{Results}

\begin{table}[t]
\centering
\footnotesize
\setlength{\tabcolsep}{5pt}
\renewcommand{\arraystretch}{1.1}
\caption{Comparison of index characteristics across datasets. UCLH (acute care) shows concise, action-oriented structure; Somerset (oncology) is more narrative; NICE (national guidance) is highly heterogeneous.}
\label{tab:index_comparison}
\begin{tabular}{lccc}
\toprule
\textbf{Statistic} & \textbf{UCLH} & \textbf{Somerset} & \textbf{NICE} \\
\midrule
Number of guidelines & 24 & 6 & 10 \\
Number of sections & 987 & 469 & 4,726 \\
Sections per guideline, mean & 41.1 & 78.2 & 472.6 \\
Sections per guideline, range & 6–74 & 48–125 & 12–849 \\
\midrule
Section length, words mean & 31.4 & 65.8 & 18.6 \\
Question length, words mean & 8.7 & 8.2 & 8.3 \\
\bottomrule
\end{tabular}
\end{table}

\paragraph{Baseline Index Construction} Applying the baseline indexing procedure (see Methods) across the three data sources yields corpora reflecting differences in each guideline style.

For the UCLH dataset, the indexing process produced 987 sections. Sections are relatively short (mean 31 words), guidelines are decomposed into an average of 41 sections (see Table~\ref{tab:index_comparison}). In the Somerset NHS Trust dataset, the index contains 469 sections across 6 guidelines (mean 78 sections per guideline). Sections are substantially longer than in UCLH (mean 66 words), reflecting a more narrative style with multiple clinical considerations rather than concise decision rules. The NICE dataset is substantially larger and more heterogeneous, comprising 4,726 sections across 10 guidelines (mean 473 sections per guideline). Sections are shorter on average (mean 19 words). Across all datasets, representative questions remain short and consistent ($\approx$ 8 words). Examples are shown in Table~\ref{tab:dataset_question_examples}. These questions capture key clinical core reasoning patterns, including diagnostic criteria, escalation decisions, and safety-critical alerts.

\begin{table}[t]
\centering
\caption{Automatic evaluation of baseline (BASE) retrieval across embedding models and representations. We compare representations based on section text, section topic and title (text only) and representations based on section text, section topic, title and representative LLM-generated question (text + question). Guideline sections are retrieved using cosine similarity between embedded user query and representations, and returned in full as answers. Metrics include ROUGE-L (R-L, sentence-level recall), word-level precision (P), recall (R), and F1, as well as medical term precision, recall, and F1. Answer length (AnsLen) is reported as mean word count (lower is better). BASE achieves consistently high recall across embeddings, but low precision and long outputs, indication non-essential content. Best results per representation are shown in \textbf{bold}, second best are underlined.}
\label{tab:base_retrieval}
\small
\begin{tabular}{lrrrrrrrr}
\toprule
\textbf{Embedding} & \textbf{Ans Len} & \textbf{R-L} & \textbf{P} & \textbf{R} & \textbf{F1} & \textbf{Term P} & \textbf{Term R} & \textbf{Term F1} \\
\midrule
\multicolumn{9}{l}{\textit{text}} \\
gemma & 190.97 & 0.3026 & 0.2117 & 0.8372 & 0.3151 & 0.2614 & 0.8191 & 0.3717 \\
nomic & 245.81 & 0.1720 & 0.1244 & 0.6518 & 0.1872 & 0.1479 & 0.6136 & 0.2166 \\
qwen3 & \underline{174.03} & \underline{0.3218} & \underline{0.2337} & \underline{0.8369} & \underline{0.3340} & \underline{0.3032} & \underline{0.8327} & \underline{0.4089} \\
\midrule
\multicolumn{9}{l}{\textit{text + question}} \\
gemma & 257.65 & 0.2248 & 0.1514 & 0.7853 & 0.2381 & 0.1857 & 0.7635 & 0.2771 \\
nomic & 255.74 & 0.2048 & 0.1366 & 0.8035 & 0.2190 & 0.1779 & 0.7860 & 0.2684 \\
qwen3 & \textbf{181.77} & \textbf{0.3332} & \textbf{0.2418} & \textbf{0.8608} & \textbf{0.3430} & \textbf{0.3083} & \textbf{0.8514} & \textbf{0.4105} \\
\bottomrule
\end{tabular}
\end{table}

\begin{table}[t]
\centering
\caption{Embedding similarity statistics for baseline retrieval. Values represent cosine similarity between the query and the top-ranked guideline section. We compare representations based on section text, section topic and title (text only) and representations based on section text, section topic, title and representative LLM-generated question (text + question). Results show overall high similarity across models, with \texttt{nomic} achieving the highest mean similarity and lowest variance, and \texttt{gemma} showing lower similarity. Best values per representation are shown in \textbf{bold}.}
\label{tab:embedding_similarity}
\small
\begin{tabular}{l l c c c}
\hline
\textbf{Representation} & \textbf{Model} & \textbf{Mean} & \textbf{Std} & \textbf{Max} \\
\hline

\multicolumn{5}{l}{\textbf{Text only}} \\

 & nomic & \textbf{0.76} & \textbf{0.05} & 0.86 \\
 & qwen3 & 0.75 & 0.06 & \textbf{0.88} \\
 & gemma & 0.65 & 0.06 & 0.83 \\

\hline

\multicolumn{5}{l}{\textbf{Text + question}} \\

 & nomic & \textbf{0.75} & \textbf{0.04} & \textbf{0.84} \\
 & qwen3 & 0.72 & 0.06 & \textbf{0.84} \\
 & gemma & 0.68 & 0.07 & 0.82 \\

\hline
\end{tabular}
\end{table}

\paragraph{Baseline retrieval limitations.}

Our BASE retrieval strategy provides an upper bound for content coverage, consistently achieving high recall across all configurations (word-level recall $=0.65$--$0.86$; Table~\ref{tab:base_retrieval}). This indicates that relevant guideline content is retrieved within the selected sections. However, precision remains substantially lower (word precision $=0.12$--$0.24$), reflecting that retrieved sections frequently include non-essential content.

BASE produces long outputs (mean 174--257 words), as entire guideline sections are returned rather than selectively extracted evidence. This reduces usability of those answers in time-critical clinical settings.

Across embedding strategies, \texttt{qwen3} achieves the strongest performance overall. In both representations, it yields the highest ROUGE-L and F1 scores, while also producing the shortest outputs among BASE variants (174 words for text-only).

Augmenting section representations with LLM-generated questions (text + question) produces mixed effects across embeddings. In particular, it is beneficial for \texttt{qwen3} and \texttt{nomic}, but decreases performance for \texttt{gemma}. Also, text + question representations tend to slightly increase output length.

All models achieve relatively high cosine similarity (mean $0.65$--$0.76$, Table~\ref{tab:embedding_similarity}), confirming that retrieval identifies semantically relevant sections. \texttt{gemma} exhibits lower mean similarity and higher variance, consistent with its weaker downstream retrieval performance.

\begin{table*}[t]
\centering
\caption{
Performance comparison on the UCLH dataset across prompting-based extraction strategies and model families. Metrics include ROUGE-L (R-L, sentence-level recall), word-level precision (P), recall (R), and F1, as well as medical term precision, recall, and F1. Answer length (AnsLen) is reported as mean word count (lower is better). Statistical significance is assessed via paired bootstrap resampling (10{,}000 draws with replacement), with 95\% confidence intervals and two-sided p-values computed from the bootstrap distribution; significance (*) indicates $p<0.05$ relative to the \textsc{COPY} baseline. \textsc{LINES} consistently achieves the highest recall among prompting methods, approaching the coverage of baseline retrieval while producing substantially shorter outputs. Bold highlights the top-performing methods per metric, second-best performance is underlined.
}
\label{tab:uclh_results}
\small
\begin{tabular}{lcccccccc}
\toprule
Method & AnsLen & R-L & P & R & F1 & Term P & Term R & Term F1 \\
\midrule
\multicolumn{9}{l}{\textbf{GPT-OSS-120B}} \\
copy   & \textbf{44.96} & \textbf{0.7537} & \textbf{0.7884} & 0.8276 & \textbf{0.7667} & \textbf{0.7979} & \underline{0.8485} & \textbf{0.7807} \\
lines  & 109.49 & 0.6231* & 0.5380 & \textbf{0.9406*} & 0.6253* & 0.5631 & \textbf{0.9485*} & 0.6540* \\
junior & 118.00 & 0.4983* & 0.4205 & 0.8745 & 0.5085* & 0.4414 & 0.8753 & 0.5298* \\
safety & \underline{45.05} & \underline{0.7082*} & \underline{0.7371} & \underline{0.8014} & \underline{0.7169*} & \underline{0.7424} & 0.8174 & \underline{0.7295*} \\
\midrule
\multicolumn{9}{l}{\textbf{Qwen2.5-32B-Instruct}} \\
copy   & \underline{39.75} & \textbf{0.7717} & \textbf{0.8199} & \underline{0.8054} & \textbf{0.7762} & \textbf{0.8230} & \underline{0.8275} & \textbf{0.7871} \\
lines  & 93.91 & 0.6800* & 0.6349 & \textbf{0.9040*} & 0.6820* & 0.6584 & \textbf{0.9128*} & 0.7036* \\
junior & 146.33 & 0.4942* & 0.4415 & 0.8317 & 0.5052* & 0.4692 & 0.8354 & 0.5335* \\
safety & \textbf{39.44} & \underline{0.6794*} & \underline{0.7163} & 0.7550* & \underline{0.6891*} & \underline{0.7341} & 0.7676 & \underline{0.7098*} \\
\bottomrule
\end{tabular}
\end{table*}

\begin{table}[t]
\centering
\caption{
Automatic evaluation of prompting strategies for GPT-OSS-120B with full Somerset NHS guideline context. Metrics include ROUGE-L (R-L, sentence-level recall), word-level precision (P), recall (R), and F1, as well as medical term precision, recall, and F1. Answer length (Len) is reported as mean word count (lower is better). 
Statistical significance is assessed via paired bootstrap resampling (10,000 draws with replacement), with 95\% confidence intervals and two-sided p-values computed from the bootstrap distribution; significance (*) indicates $p < 0.05$ relative to the \textsc{COPY} baseline.
Results highlight a clear precision–recall trade-off. \textsc{Junior} achieves the highest recall. Bold indicates the best-performing method per metric, and second-best values are underlined.
}
\label{tab:somerset_full}
\small
\begin{tabular}{lrrrrrrrr}
\toprule
\textbf{Method} & \textbf{Len} & \textbf{R-L} & \textbf{P} & \textbf{R} & \textbf{F1} & \textbf{Term P} & \textbf{Term R} & \textbf{Term F1} \\
\midrule
copy   & \underline{41.30} & \textbf{0.4051} & \textbf{0.4157} & 0.5265 & \textbf{0.4287} & \textbf{0.4332} & 0.5188 & \textbf{0.4432} \\
lines  & 147.42 & 0.2695* & 0.2311 & \underline{0.5690} & 0.2892* & 0.2944 & \underline{0.5541} & 0.3430 \\
junior & 222.92 & 0.1527* & 0.1170 & \textbf{0.6386}* & 0.1670* & 0.1475 & \textbf{0.6220}* & 0.2121* \\
safety & \textbf{34.86} & \underline{0.3922} & \underline{0.4117} & 0.4927 & \underline{0.4170} & \underline{0.4273} & 0.4967 & \underline{0.4296} \\
\bottomrule
\end{tabular}
\end{table}

\begin{table*}[t]
\centering
\caption{
Automatic evaluation of prompting strategies across model families on full NICE guidelines. Metrics include ROUGE-L (R-L, sentence-level recall), word-level precision (P), recall (R), and F1, as well as medical term precision, recall, and F1. Answer length (Len) is reported as mean word count (lower is better). Statistical significance is assessed via paired bootstrap resampling (10,000 draws with replacement), with 95\% confidence intervals and two-sided p-values computed from the bootstrap distribution; significance (*) indicates $p < 0.05$ relative to the \textsc{COPY} baseline. In this setting with very long guidelines (over 5,000 words), \textsc{Junior} consistently achieves the highest recall across models (e.g., word recall up to $\sim$0.94), with statistically significant improvements over \textsc{COPY}, while maintaining shorter outputs than \textsc{LINES}. Bold indicates the best-performing method per metric, and second-best values are underlined.
}
\label{tab:nice_full}
\small
\begin{tabular}{lrrrrrrrr}
\toprule
\textbf{Method} & \textbf{Len} & \textbf{R-L} & \textbf{P} & \textbf{R} & \textbf{F1} & \textbf{Term P} & \textbf{Term R} & \textbf{Term F1} \\
\midrule
\multicolumn{9}{l}{\textit{DeepSeek-V3}} \\
copy   & \textbf{65.50} & \textbf{0.6360} & \textbf{0.6411} & 0.8634             & \textbf{0.6944} & \textbf{0.6844} & 0.9044              & \textbf{0.7419} \\
lines  & 222.76            & 0.3446*         & 0.2776          & \underline{0.8883} & 0.3735*         & 0.3641          & \underline{0.9073}  & 0.4700* \\
junior & 147.19            & 0.3443*         & 0.2470          & \textbf{0.8963}*   & 0.3609*         & 0.3399          & \textbf{0.9597}*    & 0.4703* \\
safety & \underline{77.73}    & \underline{0.4979}* & \underline{0.4207} & 0.8578        & \underline{0.5234}* & \underline{0.4856} & 0.9128         & \underline{0.5990}* \\
\midrule
\multicolumn{9}{l}{\textit{GPT-5-mini}} \\
copy   & \textbf{68.66} & \textbf{0.5964} & \textbf{0.5856} & 0.8911             & \textbf{0.6558} & \textbf{0.6280} & 0.9184              & \textbf{0.7027} \\
lines  & 440.75            & 0.2111*         & 0.1472          & \textbf{0.9433}* & 0.2279*        & 0.2107          & 0.9435* & 0.3110* \\
junior & 193.01            & 0.3199*         & 0.2317          & \underline{0.9155}*            & 0.3340*         & 0.2795          & \textbf{0.9718}*    & 0.4411* \\
safety & \underline{91.16}    & \underline{0.4651}* & \underline{0.3682} & 0.8774        & \underline{0.4848}* & \underline{0.4668} & \underline{0.9615}*        & \underline{0.5932}* \\
\midrule
\multicolumn{9}{l}{\textit{Qwen3-235B}} \\
copy   & \textbf{46.34} & \textbf{0.6296} & \textbf{0.6504} & 0.8250             & \textbf{0.6938} & \textbf{0.6832} & \underline{0.8467}  & \textbf{0.7285} \\
lines  & 225.03            & 0.3442*         & 0.2919          & \underline{0.8524} & 0.3850*         & 0.3435          & 0.8515              & 0.4412* \\
junior & 96.14             & \underline{0.4840}* & \underline{0.4187} & \textbf{0.9080}* & \underline{0.5228}* & \underline{0.4997}* & \textbf{0.9457}* & \underline{0.6115}* \\
safety & \underline{65.88}    & 0.4976*         & 0.4593          & 0.8029             & 0.5438*         & 0.5021          & 0.8273              & 0.5937* \\
\bottomrule
\end{tabular}
\end{table*}

\paragraph{Prompting-based extraction strategies.}

Across all datasets and model families, prompting-based extraction substantially improves over baseline retrieval, not by retrieving more information, but by selecting it more precisely. Among these approaches for UCLH, \textsc{LINES} consistently emerges as the most reliable strategy for preserving clinically relevant content (Table~\ref{tab:uclh_results}).

Unlike \textsc{COPY}, which serves as a baseline for verbatim extraction, \textsc{LINES} explicitly enforces line-level selection of evidence. This constraint proves critical. Across settings, \textsc{LINES} achieves the highest recall across both word-level and entity-level metrics (e.g., word recall $0.94$ and NER recall $0.95$ for GPT-OSS; $0.90$ and $0.91$ for Qwen), with statistically significant improvements over \textsc{COPY}. In safety-critical settings, where missing information may have direct consequences, this recall advantage is decisive. \textsc{LINES} therefore provides the most effective balance of completeness and usability.

\textsc{COPY} provides a strong precision-oriented baseline.
\textsc{Safety} behaves very similarly to \textsc{COPY}, achieving  precision close to \textsc{COPY} across settings, while producing outputs of similar length. \textsc{Junior} improves recall relative to other methods, but consistently fall short of \textsc{LINES} while producing longer and less precise responses.

\begin{figure}[t]
\centering
\includegraphics[width=0.8\textwidth]{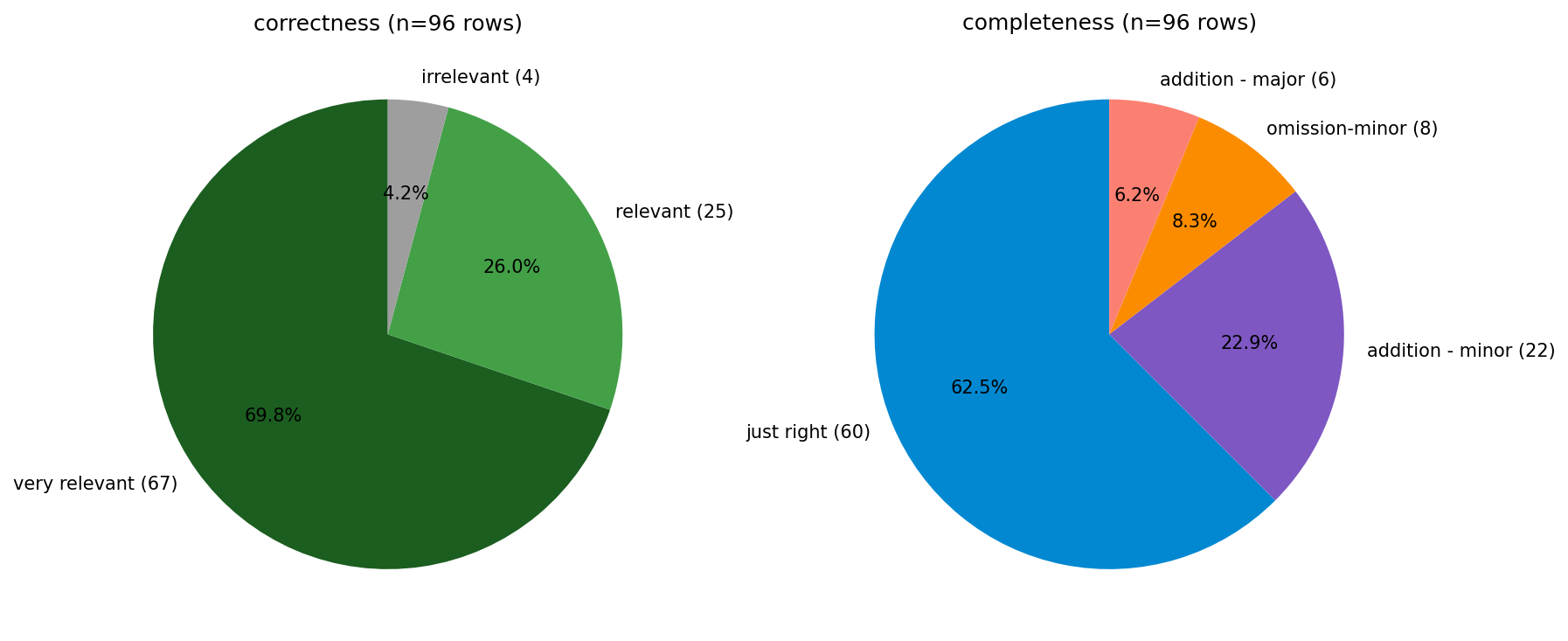}
\caption{Clinician evaluation of LINES outputs. Distribution of relevance (left) and completeness (right) ratings across 96 clinical questions. LINES achieves high clinical relevance (95.8\% at least relevant) and the highest proportion of ``just right'' answers (62.5\%), indicating strong alignment with reference guidance. No major omissions were observed, suggesting that LINES reliably preserves critical clinical content.}
\label{fig:lines_eval}
\end{figure}

\begin{figure}[t]
\centering
\includegraphics[width=0.8\textwidth]{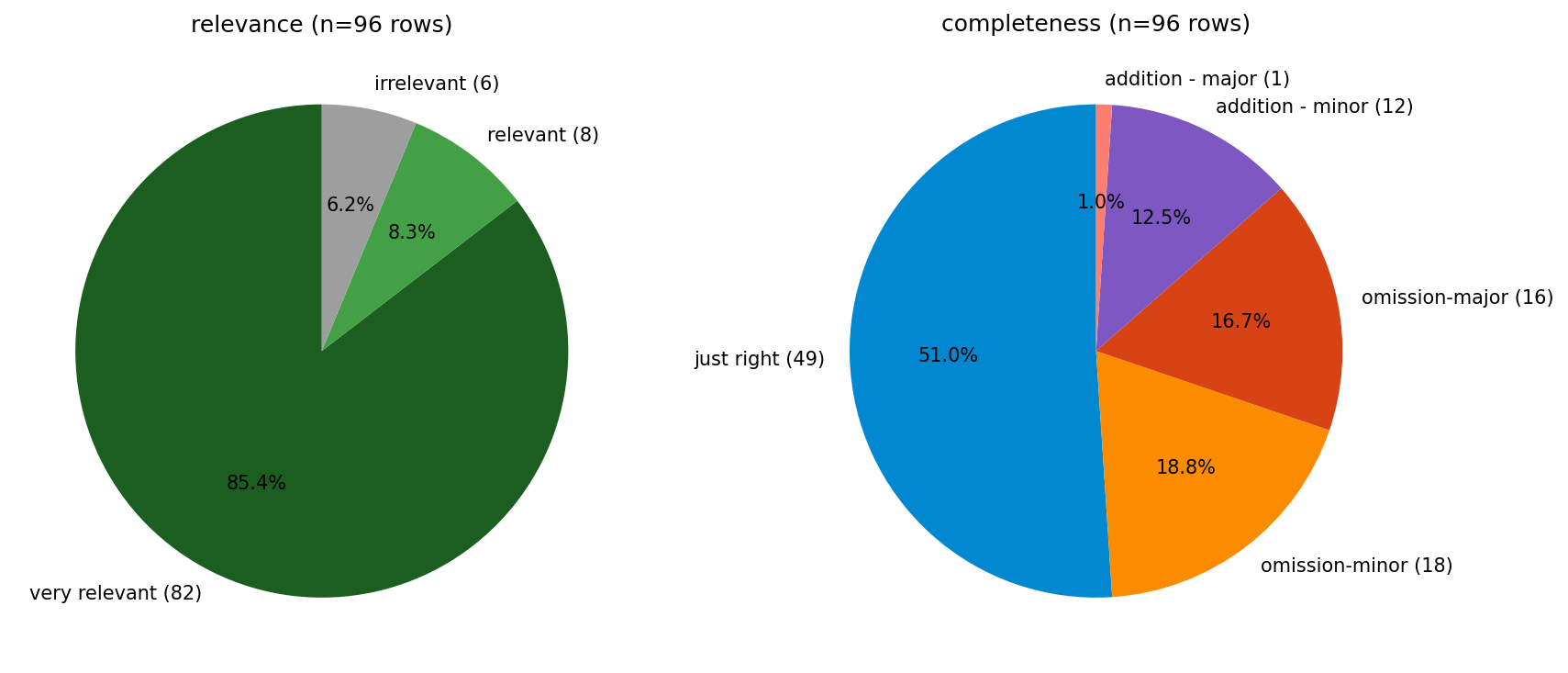}
\caption{Clinician evaluation of Safety outputs. Distribution of relevance (left) and completeness (right) ratings across 96 clinical questions. Safety produces highly relevant outputs (93.7\% at least relevant), with a higher proportion of ``very relevant'' ratings compared to LINES, reflecting concise and focused responses. The strategy presents increased omissions (35.5\%, including 16.7\% major omissions).}
\label{fig:safety_eval}
\end{figure}

\paragraph{Extraction behaviour analysis.}

Table~\ref{tab:extraction_examples} provides qualitative insight into how different strategies work. For example, for Pericarditis Q10, the reference answer specifies three key elements: chest radiograph, transthoracic echocardiogram, and cardiology input. Both LINES and COPY recover all elements, while Safety keeps only the mimimum subset. 

For Q11 Pericarditis, LINES most closely matches the reference answer with all key elements. COPY also achieves high coverage but introduces minor omission of the cardiology follow-up. In contrast, the Safety strategy produces a more concise output, prioritising core treatment steps but omits important supporting details, notably PPI co-prescription, GI side-effect management, and cardiology follow-up. These trends are consistent across both GPT-OSS and Qwen models.

Qualitative analysis of JuniorDoc (Table~\ref{tab:junior_examples_revised}) shows that Agent-2 effectively consolidates key guideline elements. However, additions of non-essential contextual information and omission of relevant details are frequent. For example, in AF, Agent-2 captures the core decision rule for cardioversion (onset $<$48\,h with no precipitant), but introduces additional context (broader rhythm control pathways).

\paragraph{Performance of extraction strategies for longer documents and frontier models.}

Overall, the performance of our extraction strategies depends strongly on document length and retrieval setting. When applied to full, long-form guidelines (Somerset and NICE), the JuniorDoc multi-agent approach becomes increasingly competitive, and in most of the cases achieves the highest recall across models (Tables~\ref{tab:somerset_full} and \ref{tab:nice_full}). For example, on full NICE guidelines (often exceeding 5,000 words), JuniorDoc consistently achieves the strongest word-level and medical term recall across DeepSeek, GPT-5-mini, and Qwen3, with statistically significant improvements over \textsc{COPY}. 

This behaviour may be attributed to the prompt pipeline: agent~1 generates a response draft aligned with clinicians' ideas about the problem, while agent~2 retrieves supporting guideline evidence. In long documents, this enables the model to pinpoint evidence that would be difficult to capture in this long context. As a result, JuniorDoc is well-suited to large guidelines such as Somerset ($\sim$ 3,000 words) and NICE ($\sim$ 5,000 words).

The pattern reverses when retrieval is constrained. When guideline sections are pre-selected (e.g., via cumulative similarity mass thresholding), \textsc{LINES} consistently achieves the highest recall across models while maintaining moderate answer length (Tables~\ref{tab:somerset_sections} and \ref{tab:nice_sections}). In this setting, the search space is reduced to a focused subset of relevant content (e.g., for NICE from $\sim$5,000 words per full guideline to $\sim$800 words per selected sections).

\paragraph{Human Evaluation}

Following strong automatic performance, we conducted a detailed human evaluation of model outputs generated by the GPT-OSS-120B model across 96 UCLH clinical questions, comparing both the LINES and Safety strategies (Figures~\ref{fig:lines_eval} and~\ref{fig:safety_eval}). 
The Safety strategy was additionally included as a controlled contrast in output length. This contrast provides a useful lens for assessing whether metric gains reflect true clinical quality.
Clinical review was performed by three clinicians (FJ, EG, NS), assessing both relevance and completeness. Relevance was rated as \textit{irrelevant}, \textit{relevant}, or \textit{very relevant}, reflecting appropriateness to the clinical scenario. Completeness captured whether clinically important information was omitted or whether non-essential content was included.

\textbf{Relevance.}
Both strategies produced highly clinically appropriate outputs. For LINES, 69.8\% of responses were rated as \textit{very relevant} and 26.0\% as \textit{relevant}, with only 4.2\% classified as \textit{irrelevant} (4 answers). Safety achieved even higher top-tier relevance, with 85.4\% of outputs rated as \textit{very relevant} and 8.3\% as \textit{relevant}, though a slightly higher proportion (6.2\%) were judged \textit{irrelevant}. 

\textbf{Completeness.}
Differences appear more clearly in completeness. LINES produced the highest proportion of \textit{just right} answers (62.5\%), with relatively few omissions (8.3\% minor, 0\% major), but more frequent additions (22.9\% minor, 6.2\% major), reflecting its tendency to include additional contextual detail. In contrast, Safety produced fewer additions (12.5\% minor, 1.0\% major) and a lower proportion of excessive content overall, but at the cost of increased omissions (18.8\% minor, 16.7\% major). Only 51.0\% of Safety outputs were rated as \textit{just right}.
 
 Overall, LINES prioritises completeness and coverage, rarely omitting critical information but sometimes including non-essential content. Safety enforces brevity and reduces over-generation, but more frequently omits clinically relevant details, including in some cases information that may be important for safe decision-making.

\textbf{Illustrative examples.}
The pancreatitis example (see Table~\ref{tab:minor_examples}) demonstrates a minor addition for LINES. The response appropriately prioritises urgent referral to ICU outreach, including clear clinical criteria to support severity assessment and escalation decisions. It also provides relevant contact pathways, including the general medical registrar, ICU outreach services, hepatobiliary specialists, and dietetics, all of which are appropriate for ongoing care. The inclusion of additional sections on ongoing management and imaging is not directly required to answer the referral question. However, this information remains clinically relevant and does not detract from decision-making.

The ureteric colic example illustrates a minor omission. The response provides a clear and clinically useful set of discharge criteria that are easy to interpret and apply. One element of the guideline, relating to stone size and anatomical location, is not explicitly included and would require reference to an associated table. This omission does not significantly affect clinical decision-making, as it can be readily clarified if needed and does not compromise patient safety.

\textbf{Inter-annotator agreement.}
Inter-annotator agreement was assessed on a subset of 12 shared questions (Pericarditis, Acute Kidney Injury, and Atrial Fibrillation). For relevance, unanimous agreement was observed in 3 cases, with the remaining 9 showing agreement between two annotators. Disagreements were limited to adjacent categories (e.g. \textit{relevant} vs \textit{very relevant}). For completeness, full agreement was observed in 6 cases, with the remaining 6 showing partial agreement. Disagreements again occurred primarily between neighbouring categories, most commonly \textit{just right} and \textit{addition--minor}. Majority voting was used to assign final labels.

These findings confirm that the \textsc{LINES} strategy provides a strong mechanism for clinically evidence extraction. The consistently high relevance (95.8\%) and absence of major omissions indicate that \textsc{LINES} effectively preserves critical safety-relevant information. Minor omissions tend to affect peripheral details, confirming that \textsc{LINES} avoids unsafe under-specification.

\section*{Discussion}

\paragraph{Statement of key findings}

Our results demonstrate that clinically relevant information can be reliably delivered by an LLM-based system with relevant prompting. A central finding is that the effectiveness of extraction strategies depends critically on document length and retrieval setting. For shorter protocol-like guidelines, \textsc{LINES} consistently achieves the highest term recall across models while maintaining moderate answer length (+10 points increase). In this setting, the reduced search space enables precise line-level selection.

When operating over long guidelines (e.g., Somerset and NICE), the \textsc{JuniorDoc} approach becomes more effective and often achieves the highest term recall (+7 points on average). This reflects its ability to draft required evidence first and then more efficiently find it in long documents.  

These results highlight a fundamental trade-off between completeness of extractive approaches and conciseness of abstractive approaches. While \textsc{LINES} provides the highest information recall for shorter documents, draft-first approaches \textsc{JuniorDoc} are better suited to large-context settings. Strategies that re-interpret input text (e.g., \textsc{Copy} and \textsc{Safety}) improve brevity but may omit clinically relevant information, hence raising safety concerns. 

\paragraph{Strengths and limitations}

This study introduces an approach to clinical question answering in which LLMs are constrained to operate as evidence extractors rather than free-text generators. By enforcing reproduction of guideline content, the approach directly addresses one of the key risks of LLM deployment in healthcare, the risk of hallucination, while enabling transparent and auditable outputs.

A further strength is the systematic comparison of prompting strategies across different document-length regimes. This reveals that performance is not solely a function of model capacity, but of the interaction between prompting strategy and input structure. In particular, the results demonstrate that increasing architectural complexity (e.g. multi-agent pipelines) may not always be beneficial.

However, several limitations should be mentioned. First, human evaluation was conducted on a relatively small set of clinical questions, and although designed to reflect realistic scenarios, larger-scale validation is required. Second, the study focuses on text-based guidelines and does not include common tabular or visual elements (e.g. flowcharts). Extending the approach to multimodal guideline formats remains an important direction.

Third, evaluation was performed in a simulated setting rather than in real-time clinical workflows. While this enables controlled comparison, it does not fully capture real-world usage constraints. Finally, although we evaluated several models, results may further vary with different architectures and context lengths. Nevertheless, the consistency of observed patterns across models suggests that the findings are primarily driven by prompting strategy rather than model-specific effects.

\paragraph{Implications and next steps}

These findings have important implications for the deployment of LLMs in clinical settings. First, they suggest that retrieval and extraction should be jointly designed. Systems operating over large contexts may benefit from draft-based approaches such as \textsc{Junior}, whereas systems operating over shorter contexts can rely on extraction methods such as \textsc{LINES}.

Second, the results emphasise the need to explicitly manage the precision–recall trade-off. In safety-critical applications, prioritising recall may be necessary to avoid omission of clinically relevant information, even at the cost of longer outputs. Future systems may benefit from adaptive strategies that dynamically adjust verbosity based on task complexity.

Future work will focus on real-world deployment, including integration with electronic health record systems and evaluation in live clinical environments. In addition, interactive capabilities, such as clarifying missing information, may enhance both completeness and relevance.

\section*{Conclusion}

This study demonstrates that LLM-based systems can provide fast and reliable to guideline information when constrained to evidence extraction. Crucially, the optimal extraction strategy depends on the retrieval setting: line number extraction is most effective protocol-like documents, whereas draft-first approaches are advantageous for long documents.

Beyond eliminating hallucination, our findings emphasise that preserving completeness is essential for patient safety. By combining extraction, drafting and retrieval techniques, LLMs can move towards becoming trustworthy clinical assistants.

\section*{Code Availability}

The code used to reproduce the experiments and evaluation pipeline is publicly available at \url{https://github.com/julia-ive/safeLLM-extraction}.

\section*{Data Availability}

The clinical guideline data used in this study are proprietary and cannot be publicly shared due to confidentiality requirements. NICE Q\&A dataset is publicly available online: \url{https://github.com/julia-ive/guidelines_qa}

\section*{Competing interests} We have no competing interests to declare. 

\section*{Funding} Not applicable

\section*{Author Contributions Statement}

JI: Conceptualisation, Methodology, Software, Validation, Formal analysis, Investigation, Writing – Final draft preparation, Writing - Reviewing and Editing. FJ: Conceptualisation, Methodology, Validation, Formal Analysis, Investigation, Writing – Early draft preparation, Writing - Reviewing and Editing. EG: Conceptualisation, Methodology, Formal Analysis, Investigation, Writing – Early draft preparation, Writing - Reviewing and Editing. NJ: Conceptualisation, Writing - Reviewing and Editing.  CSH, PB, EC: Writing - Reviewing and Editing. RD: Conceptualisation, Methodology, Writing - Reviewing and Editing. All authors approved the manuscript. 

\section*{Appendices}

\setcounter{table}{0}
\renewcommand{\thetable}{A\arabic{table}}

\begingroup
\footnotesize
\setlength{\tabcolsep}{3pt}
\renewcommand{\arraystretch}{1.05}


\begin{longtable}{p{4cm} p{11cm}}
\caption{\textbf{Clinical scenario questions for human evaluation.} Questions grouped by acute medical topics targeting decision-critical elements including diagnosis, investigation, treatment, escalation, and discharge safety.}
\label{tab:task_questions_long} \\

\toprule
\textbf{Topic} & \textbf{Question} \\
\midrule
\endfirsthead

\toprule
\textbf{Topic} & \textbf{Question} \\
\midrule
\endhead

\midrule
\multicolumn{2}{r}{\emph{Continued on next page}} \\
\midrule
\endfoot

\bottomrule
\endlastfoot

Acute coronary syndromes & Should I ever start ACS treatment before the troponin result is available? \\
Acute coronary syndromes & If I have already sent a troponin sample, when do I send a second one? \\
Acute coronary syndromes & Do I give supplementary oxygen to all ACS patients? \\
Acute coronary syndromes & My patient was not on any anticoagulation preadmission. How long do they need to stay on fondaparinux? \\

Acute diarrhea & What key discharge advice should I give to outpatients managing acute diarrhoea at home? \\
Acute diarrhea & In an inpatient who develops new diarrhoea (Bristol 5+), what tests and actions are required? \\
Acute diarrhea & Is it safe to use anti-motility agents (e.g. loperamide) or opiates in acute diarrhoea? \\
Acute diarrhea & How do I decide between oral and IV rehydration? \\

Acute kidney injury & Which nephrotoxic medications must I stop or adjust immediately? \\
Acute kidney injury & What is a ‘full renal screen’? \\
Acute kidney injury & When should I involve ICU or renal specialist for possible RRT? \\
Acute kidney injury & How do I manage pulmonary oedema caused by AKI? \\

AF & How do I decide if cardioversion is safe based on time of onset? \\
AF & What monitoring and nursing observations are required once therapy is started? \\
AF & Which baseline investigations must I request on initial assessment? \\
AF & How do I risk stratify for bleeding with anticoagulation? \\

Cardiogenic pulmonary oedema & What oxygen saturation targets should I aim for, and when is oxygen indicated? \\
Cardiogenic pulmonary oedema & Which baseline medications should I review or stop immediately? \\
Cardiogenic pulmonary oedema & Which imaging modalities should I arrange, and how urgently? \\
Cardiogenic pulmonary oedema & What is the recommended initial IV diuretic regimen? \\

COPD & What immediate investigations and samples should I request on arrival? \\
COPD & Which patients warrant ID/Micro review? \\
COPD & What criteria mandate early ICU referral? \\
COPD & When should I consider aminophylline? \\

Decompensated liver cirrhosis & What bowel washout regimen should I start for hepatic encephalopathy? \\
Decompensated liver cirrhosis & What VTE prophylaxis is appropriate and what are contraindications? \\
Decompensated liver cirrhosis & What acute hepatic screen should I order? \\
Decompensated liver cirrhosis & Who would I call for urgent ascitic drain insertion? \\

DKA & Once stable, what maintenance fluid regimen is recommended? \\
DKA & How often should I monitor CBG/VBG and electrolytes? \\
DKA & When should I discuss with ICU for escalation? \\
DKA & Is there an order set for insulin prescribing? \\

GCA & Should I refer to rheumatology or ophthalmology? \\
GCA & Should I treat first or discuss first? \\
GCA & What symptoms suggest complicated GCA? \\
GCA & Beyond steroids, what treatment is required? \\

Hot swollen joint & What are the main causes? \\
Hot swollen joint & What investigations are required? \\
Hot swollen joint & How is septic arthritis treated? \\
Hot swollen joint & What is first-line treatment for gout? \\

Hypercalcaemia & What are the main symptoms? \\
Hypercalcaemia & What investigations beyond bloods are required? \\
Hypercalcaemia & At what level is it considered severe? \\
Hypercalcaemia & Which drugs commonly cause it? \\

Hyperkalaemia & What are the severity cut-offs? \\
Hyperkalaemia & What is pseudohyperkalaemia? \\
Hyperkalaemia & What are ECG signs? \\
Hyperkalaemia & When is cardiac monitoring required? \\

HHS & What are the diagnostic criteria? \\
HHS & What monitoring is needed? \\
HHS & What if systolic BP is below 90? \\
HHS & How do I know resolution? \\

Hypocalcaemia & What are the symptoms? \\
Hypocalcaemia & What are ECG signs? \\
Hypocalcaemia & What monitoring is required? \\
Hypocalcaemia & What if vitamin D deficient? \\

Hypokalaemia & At what level does it require treatment? \\
Hypokalaemia & When can oral replacement be used? \\
Hypokalaemia & What are ECG signs? \\
Hypokalaemia & When should concentrated potassium be used? \\

Hyponatraemia & When is it considered severe? \\
Hyponatraemia & What investigations are required? \\
Hyponatraemia & How to test for pseudohyponatraemia? \\
Hyponatraemia & How is it treated? \\

Infective endocarditis & What antibiotics should be given? \\
Infective endocarditis & What blood tests are required? \\
Infective endocarditis & What imaging should be ordered? \\
Infective endocarditis & Who should this be discussed with? \\

SAH & How is diagnosis confirmed? \\
SAH & What are the symptoms? \\
SAH & When should LP be performed? \\
SAH & How should CT-confirmed SAH be managed? \\

Pancreatitis & How is diagnosis confirmed? \\
Pancreatitis & What tests should be ordered? \\
Pancreatitis & Who should I refer to? \\
Pancreatitis & When should I escalate to ICU? \\

Pericarditis & What imaging should I request? \\
Pericarditis & What ECG changes occur? \\
Pericarditis & What is the treatment? \\
Pericarditis & Can the patient be discharged? \\

Tachycardia & What investigations should I perform? \\
Tachycardia & What are differentials for broad complex tachycardia? \\
Tachycardia & How do I escalate unstable tachycardia? \\
Tachycardia & What electrolyte targets should I aim for? \\

Ureteric colic & What tests should I perform? \\
Ureteric colic & What is the treatment? \\
Ureteric colic & When is discharge safe? \\
Ureteric colic & When should I refer to urology? \\

Upper GI bleed & How do I assess severity? \\
Upper GI bleed & When to activate major haemorrhage protocol? \\
Upper GI bleed & What additional blood products are needed? \\
Upper GI bleed & What is the dose of terlipressin? \\

\end{longtable}


\begin{longtable}{p{4cm} p{11cm}}
\caption{\textbf{Oncology guideline-based clinical questions.}}
\label{tab:oncology_questions} \\

\toprule
\textbf{Guideline} & \textbf{Question} \\
\midrule
\endfirsthead

\toprule
\textbf{Guideline} & \textbf{Question} \\
\midrule
\endhead

\bottomrule
\endlastfoot

Hypersensitivity & Irinotecan anticholinergic syndrome \\
Hypersensitivity & Drugs given in grade 2 hypersensitivity reactions \\
Hypersensitivity & How to restart chemotherapy after a reaction \\
Hypersensitivity & Oxaliplatin hypersensitivity reaction \\

Diarrhea (SACT) & Are there contraindications to DOAC? \\
Diarrhea (SACT) & Can DOAC be used in severe renal impairment? \\
Diarrhea (SACT) & Do I need monitoring on DOAC? \\
Diarrhea (SACT) & How long to treat cancer-associated VTE? \\
Diarrhea (SACT) & What if intolerant to DOACs? \\
Diarrhea (SACT) & Anticoagulation in BMI 135 kg \\
Diarrhea (SACT) & PICC-associated thrombus management \\
Diarrhea (SACT) & When to offer IVC filter \\
Diarrhea (SACT) & Apixaban dosing in cancer VTE \\
Diarrhea (SACT) & VTE with low platelets \\
Diarrhea (SACT) & Octreotide administration \\
Diarrhea (SACT) & Immunotherapy-related diarrhea management \\
Diarrhea (SACT) & Capecitabine overdose management \\
Diarrhea (SACT) & Stool cultures before loperamide? \\
Diarrhea (SACT) & Treatment for severe diarrhea \\
Diarrhea (SACT) & Investigations for SACT diarrhea \\

MSCC & Can patient be mobilised? \\
MSCC & How to contact coordinator? \\
MSCC & Management after MRI confirmation \\
MSCC & Discharge if MRI negative? \\
MSCC & LMWH prophylaxis? \\
MSCC & Steroid initiation? \\
MSCC & Clinical signs of MSCC \\
MSCC & Investigations for unknown primary \\
MSCC & First-line investigation \\
MSCC & Who to contact locally \\

Magnesium disorders & Contraindicated drugs \\
Magnesium disorders & Need for ECG \\
Magnesium disorders & Managing oral magnesium side effects \\
Magnesium disorders & Admission threshold Mg 0.45 \\
Magnesium disorders & Symptoms \\
Magnesium disorders & Causative drugs \\
Magnesium disorders & Oral supplementation \\
Magnesium disorders & Definition of severe hypomagnesaemia \\
Magnesium disorders & IV monitoring \\
Magnesium disorders & Other electrolytes \\

Neutropenic sepsis & Infection with normal neutrophils? \\
Neutropenic sepsis & Central line use \\
Neutropenic sepsis & Need for chest X-ray \\
Neutropenic sepsis & Diagnostic criteria \\
Neutropenic sepsis & Initial management \\
Neutropenic sepsis & Leg swelling post-chemo \\
Neutropenic sepsis & Antibiotics \\
Neutropenic sepsis & Escalation after failure \\
Neutropenic sepsis & Required investigations \\
Neutropenic sepsis & Neutrophil threshold \\

\end{longtable}

\endgroup

\section*{Prompt Listings}

\begin{lstlisting}[caption={Baseline Prompt: Section Subpart Extraction}, label={lst:baseline-subparts}]
Create smaller clinically coherent subparts from the paragraph and identify relevant questions.

PARAGRAPH (verbatim):
---
{paragraph}
---

Task:
- Identify 2-6 subparts that can each answer a distinct clinical question.
- Subparts may overlap and may be different lengths.
- For each subpart:
  - Provide a short title
  - Provide ONE short question (<= 16 words)
  - Provide the supporting 'text' as a verbatim span copied EXACTLY from the paragraph.

Rules:
- Copy paragraph text exactly; do not modify wording.
- Do NOT invent content not present in the paragraph.
- JSON ONLY. No markdown or commentary.

Return STRICT JSON with this schema:

{
  "subparts": [
    {
      "section_id": "P1-S1",
      "title": "string",
      "question": "<= 16 words; actionable",
      "text": "verbatim span copied EXACTLY from the paragraph"
    }
  ]
}
\end{lstlisting}

\subsection*{Topic Selection (TS)}

\begin{lstlisting}[caption={Topic Selection Prompt (TS)}, label={lst:topic-selection}]
You are selecting the most relevant clinical guideline topic.

USER QUESTION:
{user_question}

AVAILABLE GUIDELINE TOPIC IDs:
{guideline_ids}

INSTRUCTIONS:
1. Select exactly ONE guideline_id from the list above.
2. Choose the most directly relevant topic.
3. If no perfect match exists, select the closest relevant topic.
4. Do NOT invent new IDs.
5. Do NOT output explanations outside JSON.

OUTPUT FORMAT (STRICT JSON ONLY):
{
  "guideline_id": "<one of the listed IDs>",
  "rationale": "<max 1 short sentence>"
}
\end{lstlisting}

\subsection*{Line Selection Strategy (LINES)}

\begin{lstlisting}[caption={Line Selection Prompt (LINES)}, label={lst:line-selection}]
You will answer the user's question by selecting the most relevant
lines from the guideline.

GOAL:
- Select the MAXIMUM number of guideline lines that are relevant.
- When unsure whether a line is relevant, INCLUDE it rather than omit it.

GUIDELINE (with line numbers):
{numbered_guideline_text}

User question:
{user_question}

Instructions:
- At the VERY END of your response, output ONLY line numbers
  (comma-separated or ranges like 12-18).
- Do NOT include any other text after the final line numbers.
\end{lstlisting}

\subsection*{Verbatim Sentence Extraction (COPY)}

\begin{lstlisting}[caption={Verbatim Sentence Extraction Prompt (COPY)}, label={lst:copy}]
Copy the most relevant lines from the text to answer the user's question.

GUIDELINE:
{guideline_sentences}

User question:
{user_question}

Rules:
- Copy ONLY from the provided TEXT.
- Copy verbatim (exact wording).
- Prefer the minimum set of lines that fully answers the question.
- Return JSON only.

Schema:
{"copied": "verbatim lines copied from the provided text"}
\end{lstlisting}

\subsection*{JuniorDoc Strategy}

\subsubsection*{Agent 1: Draft Clinical Answer}

\begin{lstlisting}[caption={JuniorDoc Prompt: Agent 1 (Draft Clinical Answer)}, label={lst:juniordoc-agent1}]
You are Agent 1, a junior doctor answering from memory.

Provide your best approximate answer WITHOUT reading
any external documents.

Be concise and clinic-facing.

Extract actionable details whenever remembered
(e.g. thresholds, doses, durations, monitoring steps).
\end{lstlisting}

\subsubsection*{Agent 2: Guideline Evidence Extraction}

\begin{lstlisting}[caption={JuniorDoc Prompt: Agent 2 (Guideline Evidence Extraction)}, label={lst:juniordoc-agent2}]
You are Agent 2.

Inputs:
- user question
- guideline text
- Agent1 answer (for comparison only)

Task:
Select the MAXIMUM subset of FULL SENTENCES copied
VERBATIM from the guideline that either:

- SUPPORT Agent1
- ADD detail to Agent1
- CORRECT Agent1

Rules:
- Copy only from guideline sentences
- Preserve order
- Return JSON only

{"subset": ["sentence", ...]}
\end{lstlisting}

\subsection*{Minimal Safe Extraction (Safety)}

\begin{lstlisting}[caption={Minimal Safe Extraction Prompt (Safety)}, label={lst:minimal}]
Select the SMALLEST ordered subset of FULL sentences
from the guideline that answers the question.

Each sentence must have one label:

answers_question
adds_required_detail
safety_critical

Rules:
- Copy only from guideline sentences
- Preserve order
- No paraphrasing
- Return JSON

{
 "sentences": [
   {"text": "sentence", "reason": "answers_question"}
 ]
}
\end{lstlisting}

\begin{table}[t]
\centering
\footnotesize
\setlength{\tabcolsep}{6pt}
\renewcommand{\arraystretch}{1.15}
\caption{Representative questions across datasets.}
\label{tab:dataset_question_examples}
\begin{tabular}{p{3.2cm}p{3.2cm}p{7cm}}
\toprule
\textbf{Dataset} & \textbf{Topic} & \textbf{Representative Question} \\
\midrule

\textbf{UCLH} 
& Atrial Fibrillation 
& When should urgent DC cardioversion be considered in atrial fibrillation? \\

& Hyperkalaemia 
& When is urgent assessment required for hyperkalaemia? \\

& Diabetic Ketoacidosis 
& Does the patient meet DKA diagnostic criteria based on glucose, ketones, and pH? \\

\midrule

\textbf{Somerset} 
& Neutropenic Sepsis 
& What clinical features and risk factors should be assessed when managing suspected neutropenic sepsis? \\

& Chemotherapy Toxicity 
& How should chemotherapy-induced toxicity be graded and managed in oncology patients? \\

& Cancer Pain 
& What factors should guide escalation of analgesia in patients with cancer-related pain? \\

\midrule

\textbf{NICE} 
& Acute Coronary Syndrome 
& What investigations, risk stratification, and initial management steps are recommended for suspected acute coronary syndrome? \\

& Chronic Kidney Disease 
& How should CKD be diagnosed, staged, and initially managed based on eGFR and albuminuria? \\

& Stroke 
& What criteria determine urgent imaging and thrombolysis eligibility in suspected acute stroke? \\

\bottomrule
\end{tabular}
\end{table}

\begin{table*}[t]
\footnotesize
\setlength{\tabcolsep}{2.5pt}
\renewcommand{\arraystretch}{1.05}
\caption{Examples of extraction behaviour across prompting strategies (UCLH, gpt-oss:120B). The table compares reference answers with outputs generated by the LINES, COPY, and Safety strategies. LINES reconstructs answers from selected guideline lines and consistently achieves high coverage of reference content. COPY copies verbatim sentences from the guideline. Safety selects the minimal subset of sentences required to answer the question, produces concise responses but may omit important details. Original text is paraphrased.}
\label{tab:extraction_examples}

\begin{tabularx}{\textwidth}{
p{2.2cm} 
p{3.2cm} 
X 
X 
X 
X}
\toprule
Question ID & Question & Reference answer & COPY output & LINES output & Safety output \\
\midrule

Q10\_Pericarditis &
What imaging should I request if suspecting pericarditis? &
• Chest radiograph \newline
• Transthoracic echocardiogram to evaluate for pericardial effusion or tamponade \newline
• Seek cardiology input if a cardiac MRI is being considered &
REQUEST \newline
• Chest radiograph \newline
• Transthoracic echocardiogram to evaluate for pericardial effusion or tamponade \newline
• Seek cardiology input if a cardiac MRI is being considered &
• Chest radiograph \newline
• Transthoracic echocardiogram to evaluate for pericardial effusion or tamponade \newline
• Seek cardiology input if a cardiac MRI is being considered &
• Chest radiograph \newline
• Transthoracic echocardiogram to evaluate for pericardial effusion or tamponade \\

\midrule

Q11\_Pericarditis &
What is the treatment for pericarditis? &
• Ibuprofen 400 mg TDS 7–10 days, then taper \newline
• PPI (e.g., lansoprazole 30 mg OD) \newline
• Colchicine 500 µg BD (or OD if <75 kg) for 3 months \newline
• Adjust for GI side effects \newline
• Avoid routine steroids \newline
• Limit physical activity \newline
• Cardiology follow-up &
• Ibuprofen 400 mg TDS 7–10 days, then taper \newline
• PPI (e.g., lansoprazole 30 mg OD) \newline
• Colchicine 500 µg BD (or OD if <75 kg) for 3 months \newline
• Adjust for GI side effects \newline
• Avoid routine steroids \newline
• Limit physical activity &
• Ibuprofen 400 mg TDS 7–10 days, then taper \newline
• PPI (e.g., lansoprazole 30 mg OD) \newline
• Colchicine 500 µg BD (or OD if <75 kg) for 3 months \newline
• Adjust for GI side effects \newline
• Avoid routine steroids \newline
• Limit physical activity \newline
• Cardiology follow-up &
• Ibuprofen 400 mg TDS 7–10 days \newline
• Colchicine 500 µg BD (or OD if <75 kg) \newline
• Avoid routine steroids \newline
• Limit physical activity \\

\bottomrule
\end{tabularx}
\end{table*}

\begin{table*}[t]
\centering
\caption{
JuniorDoc pipeline examples (UCLH, gpt-oss:120B). The table illustrates how intermediate steps transform the original uninformed LLM answer. Comparison with reference answers highlights presence of non-essential content. Original text is paraphrased.
}
\label{tab:junior_examples_revised}
\small
\begin{tabular}{l p{3.2cm} p{3.5cm} p{3.8cm} p{3.5cm}}
\hline
\textbf{Question ID} & \textbf{Question} & \textbf{Agent-1 answer} & \textbf{Agent-2 sentences} & \textbf{Reference answer} \\
\hline

Pericarditis 
& What imaging should I request if suspecting pericarditis? 
& Echocardiogram is primary imaging. Request TTE. Consider bedside echo if unstable. Follow with formal echo if needed. 
& REQUEST Imaging: Chest radiograph. Transthoracic echocardiogram to evaluate for pericardial effusion or tamponade. Seek cardiology input if cardiac MRI considered. 
& Chest radiograph; Transthoracic echocardiogram to evaluate for pericardial effusion or tamponade; Seek cardiology input if a cardiac MRI is being considered \\

\hline

AF 
& How do I decide if cardioversion is safe based on time of onset? 
& Depends on AF duration: $<$48h safe without anticoagulation; 48h–1 week consider TEE; $>$1 week anticoagulate 3 weeks before cardioversion 
& DIAGNOSIS: onset $<$48h. RHYTHM CONTROL: cardioversion if within 48h and no precipitant. If $>$48h: rate control + anticoagulation then elective cardioversion. 
& Consider cardioversion if onset $<$48h and no clear precipitant \\

\hline
\end{tabular}
\end{table*}

\begin{table}[t]
\centering
\caption{
Automatic evaluation of prompting strategies for GPT-OSS-120B with Somerset NHS guideline sections selected up to a cumulative similarity mass threshold (0.95) using the text+question representations. Under this setting, \textsc{LINES} achieves consistently higher recall than other prompting methods (e.g., word recall $\approx$0.65), with statistically significant improvements over \textsc{COPY}, while maintaining moderate answer length. Multi-agent (JuniorDoc) further increases recall marginally but at the cost of longer outputs. Bold indicates the best-performing method per metric, and second-best values are underlined.
}
\label{tab:somerset_sections}
\small
\begin{tabular}{lrrrrrrrr}
\toprule
\textbf{Method} & \textbf{Len} & \textbf{R-L} & \textbf{P} & \textbf{R} & \textbf{F1} & \textbf{Term P} & \textbf{Term R} & \textbf{Term F1} \\
\midrule
copy   & \underline{42.50} & \textbf{0.4321} & \textbf{0.5043} & 0.5186 & \textbf{0.4642} & \textbf{0.5034} & 0.5125 & \textbf{0.4706} \\
lines  & 117.28 & 0.3406* & 0.3006 & \underline{0.6539}* & 0.3639* & 0.3755* & \textbf{0.6291}* & 0.4191* \\
junior & 142.00 & 0.2549* & 0.2133 & \textbf{0.6543}* & 0.2828* & 0.2567* & \underline{0.6372}* & 0.3228* \\
safety & \textbf{38.62} & \underline{0.3774}* & \underline{0.4661} & 0.4596* & \underline{0.4089}* & \underline{0.4786} & 0.4742 & \underline{0.4274}* \\
\bottomrule
\end{tabular}
\end{table}

\begin{table*}[t]
\centering
\caption{
Automatic evaluation of prompting strategies across model families with retrieval constrained to NICE guideline sections selected up to a cumulative similarity mass threshold (0.95). Metrics include ROUGE-L (R-L, sentence-level recall), word-level precision (P), recall (R), and F1, as well as medical term precision, recall, and F1. Answer length (Len) is reported as mean word count (lower is better). Statistical significance is assessed via paired bootstrap resampling (10,000 draws with replacement), with 95\% confidence intervals and two-sided p-values computed from the bootstrap distribution; significance (*) indicates $p < 0.05$ relative to the \textsc{COPY} baseline. Constraining retrieval substantially reduces input context (from $\sim$5{,}000 words to $\sim$835 words). \textsc{LINES} consistently achieves the highest recall among prompting methods (e.g., word recall of $\sim$0.71 across models), with statistically significant improvements over \textsc{COPY}. Bold indicates the best-performing method per metric, and second-best values are underlined.
}
\label{tab:nice_sections}
\small
\begin{tabular}{lrrrrrrrr}
\toprule
\textbf{Method} & \textbf{Len} & \textbf{R-L} & \textbf{P} & \textbf{R} & \textbf{F1} & \textbf{Term P} & \textbf{Term R} & \textbf{Term F1} \\
\midrule
\multicolumn{9}{l}{\textit{DeepSeek-V3}} \\
copy   & \underline{65.50} & \textbf{0.6360} & \textbf{0.6411} & 0.8634             & \textbf{0.6944} & \textbf{0.6591} & 0.5912              & \textbf{0.5857} \\
lines  & 222.76            & 0.3446*         & 0.2776          & \underline{0.8883} & 0.3735*         & 0.4756          & \textbf{0.7293}*    & 0.5168* \\
junior & 147.19            & 0.3443*         & 0.2470          & \textbf{0.8963}*   & 0.3609*         & 0.4910          & \underline{0.6655}* & 0.5084* \\
safety & \textbf{77.73}    & \underline{0.4979}* & \underline{0.4207} & 0.8578        & \underline{0.5234}* & \underline{0.5453}* & 0.6370*        & \underline{0.5407}* \\
\midrule
\multicolumn{9}{l}{\textit{GPT-5-mini}} \\
copy   & \underline{68.66} & \textbf{0.5964} & \textbf{0.5856} & 0.8911             & \textbf{0.6558} & \textbf{0.6062} & 0.6604              & \textbf{0.5819} \\
lines  & 440.75            & 0.2111*         & 0.1472          & \textbf{0.9433}* & 0.2279*        & 0.4584          & \textbf{0.7209}*    & 0.4974* \\
junior & 193.01            & 0.3199*         & 0.2317          & \underline{0.9155}*            & 0.3340*         & 0.4729          & \underline{0.6958}* & 0.5054* \\
safety & \textbf{91.16}    & \underline{0.4651}* & \underline{0.3682} & 0.8774        & \underline{0.4848}* & \underline{0.5703} & \underline{0.6329} & \underline{0.5520} \\
\midrule
\multicolumn{9}{l}{\textit{Qwen3-235B}} \\
copy   & \underline{46.34} & \textbf{0.6296} & \textbf{0.6504} & 0.8250             & \textbf{0.6938} & \textbf{0.6395} & 0.6320              & \textbf{0.5967} \\
lines  & 225.03            & 0.3442*         & 0.2919          & \underline{0.8524} & 0.3850*         & 0.4397          & \textbf{0.7222}*    & 0.4878* \\
junior & 96.14             & \underline{0.4840}* & \underline{0.4187} & \textbf{0.9080}* & 0.5228* & 0.5381* & \underline{0.6747}* & \underline{0.5446}* \\
safety & \textbf{65.88}    & 0.4976*         & 0.4593          & 0.8029             & \underline{0.5438}*         & \underline{0.5906}          & 0.5919              & 0.5432* \\
\bottomrule
\end{tabular}
\end{table*}

\begin{table*}[t]
\footnotesize
\setlength{\tabcolsep}{5pt}
\renewcommand{\arraystretch}{1.15}
\caption{Examples illustrating minor addition and minor omission in LLM outputs, with clinical interpretation. All examples are paraphrased.}
\label{tab:minor_examples}

\begin{tabularx}{\textwidth}{
p{2.5cm}
p{4.5cm}
X
}
\toprule
\textbf{Type} & \textbf{Topic / Question} & \textbf{LLM Answer (verbatim)} \\
\midrule

\textbf{Minor addition} &
\textbf{Topic:} Pancreatitis \newline
\textbf{Question:} I have a patient with pancreatitis, who can I refer them to? &
\textit{``ICU REFERRAL Urgent discussion with ICU outreach SpR (via mobile)}
\begin{itemize}
\item \textit{Ongoing clinical concern or deterioration}
\item \textit{Airway not maintained (call anaesthetic SpR)}
\item \textit{GCS $\leq$ 12}
\item \textit{Respiratory distress/failure}
\item \textit{Haemodynamic instability not responding to fluids}
\item \textit{Severe electrolyte / acid-base disturbance}
\item \textit{All pancreatitis meeting `severe' Atlanta criteria}
\end{itemize}
\textit{Advice, reviews and referrals including DMR, ICU outreach, PERRT, anaesthetics, HPB and dietetics. Further sections include ongoing management and imaging guidance.''}
\\

\midrule

\textbf{Minor omission} &
\textbf{Topic:} Ureteric colic \newline
\textbf{Question:} When is it safe to discharge a patient with ureteric colic? &
\textit{``DISPOSITION Consider discharge if ALL of the following are met:}
\begin{itemize}
\item \textit{No evidence of sepsis}
\item \textit{No acute kidney injury}
\item \textit{Pain controlled}
\item \textit{Stone likely to pass spontaneously}
\item \textit{No alternative serious pathology}
\end{itemize}
\textit{Discharged patients: refer to Endourology and Stone clinic via EPIC.''}
\\
\bottomrule
\end{tabularx}
\end{table*}

\bibliography{sample}

\end{document}